\documentclass{article}

\usepackage[utf8]{inputenc} 
\usepackage[T1]{fontenc}    
\usepackage[colorlinks,citecolor=blue,linkcolor=blue]{hyperref}       
\usepackage{url}            
\usepackage{booktabs}       
\usepackage{amsfonts}       
\usepackage{nicefrac}       
\usepackage{microtype}      
\usepackage{verbatim}
\usepackage[svgnames]{xcolor}
\usepackage{amsmath,amssymb,bm}
\usepackage{algorithm}
\usepackage{algpseudocode}
\usepackage{graphicx}
\usepackage{subcaption}
\usepackage{tikz}
\usetikzlibrary{shapes,arrows,arrows.meta}
\tikzset{%
  >={Latex[width=2mm,length=2mm]},
            base/.style = {rectangle, rounded corners, draw=black,
                           minimum width=1.5cm, minimum height=1cm,
                           text centered, font=\sffamily},
  activityStarts/.style = {base, fill=blue!30},
    InExtractPortion/.style = {base, fill=blue!30},
       InfoAggregator/.style = {base, fill=red!30},
    ColGen/.style = {base, fill=green!10},
    Storage/.style = {base, fill=brown!30},
    LPSolve/.style = {base, fill=violet!30},
    ILPSolve/.style = {base, fill=red!30},
    Output/.style = {base, fill=gray!30},
       startstop/.style = {base, fill=red!30},
    activityRuns/.style = {base, fill=green!30},
         process/.style = {base, minimum width=2.5cm, fill=orange!15,
                           font=\ttfamily},
}
\pgfdeclarelayer{edgelayer}
\pgfdeclarelayer{nodelayer}
\pgfsetlayers{edgelayer,nodelayer,main}

\tikzstyle{vertex_default}=[circle,fill=White,draw=Black,text width=0.1cm]
\tikzstyle{tree_edge}=[thick,draw=Green]
\tikzstyle{add_edge}=[thick,draw=Red]
\tikzstyle{local_edge}=[thick,draw=Cyan]

\begin{document}
%
\title{Accelerating Message Passing for MAP with Benders Decomposition}
\author{
Julian Yarkony\\
Experian Data Lab. \\
\texttt{julian.e.yarkony@gmail.com} 
\and
Shaofei Wang\\
Baidu Inc. \\
\texttt{sfwang0928@gmail.com} 
}
\maketitle
\begin{abstract}
We introduce a novel mechanism to tighten the local polytope relaxation for MAP inference in Markov random fields with low state space variables.  We consider a surjection of the variables to a set of hyper-variables and apply the local polytope relaxation over these hyper-variables.  The state space of each individual hyper-variable is constructed to be enumerable while the vector product of pairs is not  easily enumerable making message passing inference intractable.

To circumvent the difficulty of enumerating the vector product of state spaces of hyper-variables we introduce a novel Benders decomposition approach.  This produces an upper envelope describing the message constructed from affine functions of the individual variables that compose the hyper-variable receiving the message. The envelope is tight at the minimizers which are shared by the true message.   Benders rows are constructed to be Pareto optimal and are generated using an efficient procedure targeted for binary problems.
\end{abstract}
\section{Introduction}
\label{sec:intro}
Linear Programming (LP) relaxations are powerful tools for finding the most probable (MAP) configuration in Markov random fields (MRF).  The most popular LP relaxation, called the local polytope relaxation \cite{kolmogorov2006convergent,wainwright2005map,globerson2008fixing,yarkony2010covering,sontag2011introduction}, is both too loose \cite{sontag} to be of use on many real problems and too computationally demanding to be solved exactly using simplex or interior point methods \cite{yanover2006linear}.  This motivates the use of coordinate updates in the Lagrangian dual, which are commonly called ``message passing"  or fixed point updates.  These updates can be applied jointly with tightening the local polytope relaxation in a cutting plane manner \cite{sontag,komodakis2008beyond}.  
%

We solve the local polytope relaxation over a graph corresponding to a surjection of the variables to sets called hyper-variables. Hyper-variables are constructed so that  iterating through the state space of each hyper-variable is feasible.  However it need not be the case that iterating through the vector product of the state spaces of pairs of hyper-variables is feasible.  For example consider a pair of hyper-variables each corresponding  to fifteen binary variables.  The state spaces of the individual hyper-variables are of the enumerable size $2^{15}$ while the corresponding vector product is of intractable size $2^{30}$. Therefor traditional message passing approaches  can not be applied as they rely on enumerating the vector product of the state spaces of hyper-variables.

In this paper we propose a Benders decomposition \cite{benders1962partitioning,geoffrion1974multicommodity,costa2005survey,wang2017exploiting} approach for computing the lower bounds on the true min-marginals needed for message passing. These lower bounds share common minimizers with the true min-marginals and the lower bound is tight at these points.  We use these lower bounds in place of the true min-marginals during message passing.   Our procedure is  guaranteed to converge to a fixed point of the Lagrangian dual problem of the local polytope relaxation.  

Our paper should be understood  in relation to \cite{wang2017efficient} which introduces nested Benders decomposition \cite{birge1985decomposition} to achieve efficient dynamic programming exact inference in high tree width binary MRFs.  The core methodological idea is to exploit the fact that the high tree width MRF is composed of low state space variables which is achieved via the nested Benders decomposition.   Our work can be understood as extending the approach of \cite{wang2017efficient} to permit message passing inference in arbitrary MRFs. 

Our work can be contrasted with its most key competitor \cite{sontag} which is motivated by protein design problems which have large state space for individual variables.  In contrast we are motivated by problems in which the states spaces of individual variables tend to be small or binary allowing for efficient specialized inference inspired by \cite{wang2017efficient}.  We now summarize the approach of \cite{sontag} which is called max-product linear programming plus cycle pursuit (MPLPCP).  

 MPLPCP alternates between message passing inference, which is applied initially over the local polytope relaxation, and adding primal constraints/dual variables.  When a fixed point is reached MPLPCP carefully selects an additional primal constraint from a bank of such constraints that is used to circumvent the fixed point and potentially tighten the relaxation.  This process iterates until the MAP is provably identified or the bank of such constraints is exhausted at a fixed point.  MPLPCP benefits from two key properties (1)  tightening the relaxation does not require restarting inference and (2) the introduction of cutting planes in the primal allows for fixed points in the dual to be bypassed.  In \cite{sontag} constraints are only produced over triples of variables and the selection of triples to add is a key bottleneck of inference\cite{batra2011tighter}.
%
%
\subsection{Outline}
We outline our paper as follows.  First in Section \ref{sec:LP-MAP} we formalize the problem of MAP inference and a corresponding message passing inference formulation \cite{globerson2008fixing}. Next in Section \ref{BendApproach} we apply Benders decomposition to produce lower bounds on the min-marginals used in message passing inference.  
Then in Section \ref{exper} we apply our approach to binary MAP inference problems that occur in multi-person pose estimation\cite{deepcut2,deepcut1,mpiiBenchmark,wang2017multi}.  Finally we conclude and  discuss extensions in Section \ref{conc}.
\section{Linear Program Relaxation for MAP Inference}
\label{sec:LP-MAP}
Consider a directed graph $G = (V,E)$ with $n$ vertices, where there is a bijection of vertices to $n$ variables $x = \{ x_1, \dots, x_n \}$.  We map a joint configuration of the variables to a cost using cost terms  $\theta_i(x_i)$ for vertex $i \in V$ and $\theta_{ij}(x_i,
x_j)$ for edge $ij \in E$, respectively
\footnote{
\label{myfoot}
Without loss of generality lets assume that $\theta$ is non-positive.  For a given problem this is achieved by subtracting the largest term present in each table $\theta_{ij}(x_i,x_j)$  and adding it to the objective.

For binary pairwise problems there is a conversion in which each pairwise table has at most  one non-zero element and that this element is non-positive.  (See Appendix \ref{binarprop})
}.  We define a function over the
variables $x$ as:
\begin{align}
\label{eqn:cost_func_binary}
f(x; \theta) = \sum_{i \in V} \theta_i(x_i) + \sum_{(ij) \in E} \theta_{ij}(x_i,x_j)
\end{align}
The MAP problem is then defined as finding an assignment that
minimizes the function $f(x; \theta)$:
\begin{align}
\label{eqn:MAP_binary}
\min_{x} \quad f(x; \theta)
\end{align}
A standard formulation of inference is the following integer linear program (ILP) which we express using over-complete representation $\mu_{ij}(x_i,x_j) \in \{0,1\}$ and $\mu_i(x_i) \{0,1\}$.  Here $\mu_{i}(x_i)$ and $\mu_{ij}(x_i,x_j)$ describe a configuration of variable $x_i$ and pair $x_i,x_j$ respectively.
\begin{align}
\label{ILPVer}
\min_{\substack{\mu \geq 0}} & \sum_{\substack{i \in V\\ x_i}} \theta_i(x_i)\mu_{i}(x_i) + \sum_{\substack{(ij) \in E \\ x_i \\ x_j}} \theta_{ij}(x_i,x_j)\mu_{ij}(x_i,x_j)\\
\nonumber \text{s.t} \quad & \sum_{x_j}\mu_{ij}(x_i,x_j) = \mu_i(x_i) \quad \forall [(ij) \in E,x_i]\\
\nonumber & \sum_{x_i}\mu_{ij}(x_i,x_j) = \mu_j(x_j) \quad \forall [(ij) \in E,x_j]\\
\nonumber & \sum_{x_i}\mu_{i}(x_i)=1 \quad \forall i \in V\\
\nonumber & \sum_{x_i,x_j}\mu_{ij}(x_i,x_j)=1  \quad \forall (ij) \in E \\
\nonumber & \mu_{i}(x_i) \in \{0,1\}  \quad \forall [i \in V,x_i]\\
\nonumber & \mu_{ij}(x_i,x_j) \in \{0,1\} \quad \forall [(ij) \in E,x_i,x_j ]
\end{align}
By relaxing the integrality constraints in Eq \ref{ILPVer} we recover a known relaxation called the local polytope relaxation \cite{sontag}.
\subsection{Tightening the Local Polytope Relaxation}
\label{tighterByBender}
%
In this sub-section we consider a tighter relaxation than the local polytope over $x$ by mapping $x$ to a new space and solving the local polytope relaxation in that space.  Consider a surjection of $\{1,2...n\}$  to $\mathcal{Y}=[\mathcal{Y}_1,\mathcal{Y}_2...\mathcal{Y}_m]$ indexed by $p$ where m<n.  Each  $\mathcal{Y}_p$ is associated with a variable  $y_p$ (which we refer to as a hyper-variable) that describes the state of all variables associated with $\mathcal{Y}_p$.  We use $x_i^{y_p}$ to denote the state of variable $x_i$ associated with $y_p$ for $i \in \mathcal{Y}_p$.  

Consider a directed graph $(\mathcal{V},\mathcal{E})$ where there is a bijection of vertices to members of $\mathcal{Y}$.  There is an edge between $p \in \mathcal{V},q\in \mathcal{V}$ in $\mathcal{E}$ if $p<q$ and there is an edge $(ij) \in E$  such that either $(i\in \mathcal{Y}_p, j \in \mathcal{Y}_q)$ or $(j\in \mathcal{Y}_p, i \in \mathcal{Y}_q)$.  
We rewrite Eq \ref{eqn:cost_func_binary} over $(\mathcal{V},\mathcal{E})$ below using  $\phi$ to aggregate the cost terms $\theta$.
\begin{align}
\label{eqn:cost_func_binary}
f(y; \phi) &= \sum_{p \in \mathcal{V}} \phi_p(y_p) + \sum_{(pq) \in \mathcal{E}} \phi_{pq}(y_p,y_q) \\
\nonumber \phi_p(y_p) &= \sum_{i \in \mathcal{Y}_p }\theta_i(x^{y_p}_i)+\sum_{\substack{i \in \mathcal{Y}_p\\ k \in \mathcal{Y}_p\\ (ik) \in E} }\theta_{ik}(x^{y_p}_i,x^{y_p}_k)\\
\nonumber \phi_{pq}(y_p,y_q) &= \sum_{\substack{i \in \mathcal{Y}_p\\ j \in \mathcal{Y}_q\\ (ij) \in E}}\theta_{ij}(x^{y_p}_i,x^{y_q}_j)+\sum_{\substack{i \in \mathcal{Y}_p\\ j \in \mathcal{Y}_q\\ (ji) \in E}}\theta_{ji}(x^{y_q}_j,x^{y_p}_i)
\end{align}
We rewrite Eq \ref{eqn:MAP_binary} in terms of $y$ and $\phi$ below.  
%
\begin{align}
\label{eqn:mapGroup}
\mbox{Eq }\ref{eqn:MAP_binary}=\min_{y} \quad f(y,\phi)
\end{align}
The local polytope relaxation over Eq\ref{eqn:mapGroup} may tighten a loose local polytope relaxation over $x$.  As a trivial example consider that there is only one set in $\mathcal{Y}$.  In that case the relaxation over $y$ is tight by definition.  Similarly if $\mathcal{E}$ forms a tree then the relaxation over $y$ is tight since the local polytope relaxation is known to be tight for tree structured graphs \cite{sontag2011introduction}.
\subsection{Message Passing Inference over the Local Polytope Relaxation}
%
%
We write the dual form of the local polytope relaxation over $(\mathcal{V},\mathcal{E})$ below using real valued dual variables $\lambda_{pq \rightarrow p}(y_p), \lambda_{pq \rightarrow q}(y_q),\lambda_p$ and $\lambda_{pq}$. 
\begin{align}
\label{eqn:ILPDualForm}
\max_{\lambda} & \sum_{p \in \mathcal{V}} \lambda_p +\sum_{(pq) \in \mathcal{E}} \lambda_{pq} \\
\nonumber \text{s.t.} \quad & \phi_p(y_p) + \sum_{\substack{q \in \mathcal{V} \\ (pq) \in \mathcal{E}}} \lambda_{pq \rightarrow p}(y_p) + \sum_{\substack{q \in \mathcal{V} \\ (qp) \in \mathcal{E}}} \lambda_{qp \rightarrow p}(y_p) \geq \lambda_p \quad \forall [ p \in \mathcal{V},y_p ] \\
\nonumber & \phi_{pq}(y_p,y_q) - \lambda_{pq \rightarrow p}(y_p) - \lambda_{pq \rightarrow q}(y_q) \geq \lambda_{pq} \quad \forall [(pq) \in \mathcal{E}, y_p, y_q]
\end{align}
In Alg \ref{MPBasic} we display the common way of optimizing Eq \ref{eqn:ILPDualForm} which iterates over $p \in \mathcal{V}$ and exactly optimizes all $\lambda$ terms in which $p$ is an index leaving all others fixed \cite{sontag2011introduction}. For ease of reading the remainder of this section and with some abuse of notation, during the inner loop operation over $p$ we flip $(qp)$ to $(pq)$ for any $(qp) \in \mathcal{E}$.
%
%
%
%
\begin{figure}
\begin{algorithm}[H]
 \caption{Basic Message Passing Algorithm}
\begin{algorithmic} 
\State Set $\lambda_{pq\rightarrow q}(y_q) = \lambda_{pq\rightarrow p}(y_p) = 0$ for all $(pq),y_p,y_q$
\Repeat
\State $\Delta \leftarrow \sum_{p \in \mathcal{V}} \lambda_p +\sum_{(pq) \in \mathcal{E}}\lambda_{pq}$
\For{$p \in \mathcal{V}$}
\For{$q \in \mathcal{V}, \text{s.t.} \ (pq) \in \mathcal{E},y_p$}
\State $\nu_{pq}(y_p) \leftarrow \min_{y_q}\phi_{pq}(y_p,y_q) -\lambda_{pq \rightarrow q}(y_q)$
\EndFor
\For{$y_p$}
\State $\nu_p(y_p) \leftarrow \frac{1}{1+\sum_{q \in \mathcal{V}} [(pq) \in \mathcal{E}]} (\phi_p(y_p)+\sum_{q; (pq) \in \mathcal{E}} \nu_{pq}(y_p))$
\EndFor
\For{$q \in \mathcal{V}, \text{s.t.} \ (pq) \in \mathcal{E},y_p$}
\State $\lambda_{pq \rightarrow p}(y_p) \leftarrow \nu_{pq}(y_p) - \nu_{p}(y_p)$
\EndFor
\State $ \lambda_p \leftarrow \min_{y_p} \phi_p(y_p)+\sum_{q, (pq) \in \mathcal{E}} \lambda_{pq \rightarrow p}(y_p)$ 
\For{$q \in \mathcal{V}, \text{s.t.} \ (pq) \in \mathcal{E}$}
\State $\lambda_{pq} \leftarrow \min_{y_p,y_q}\phi_{pq}(y_p,y_q) - \lambda_{pq \rightarrow p}(y_p) - \lambda_{pq \rightarrow q}(y_q) $
\EndFor
\EndFor
 \Until{ $\Delta= \sum_{p \in \mathcal{V}} \lambda_p +\sum_{(pq) \in \mathcal{E}}\lambda_{pq}$ }
\end{algorithmic}
\label{MPBasic}
\end{algorithm}
\caption{  We describe the standard message passing algorithm for MAP inference \cite{sontag2011introduction}.  We use $\Delta$ to indicate  the lower bound prior to a given iteration of message passing.  For ease of reading with some abuse of notation, during the inner loop operation over $p$ we flip $(qp)$ to $(pq)$ for any $(qp) \in \mathcal{E}$. }
\end{figure}

We now consider the increase in the objective at a given iterate over $p$.  We use helper term $\nu_{pq}(y_p)$ defined below which is commonly referred to as a ``min-marginal" \cite{sontag2011introduction}.  
\begin{align}
\nu_{pq}(y_p)=\min_{y_q}\phi_{pq}(y_p,y_q) -\lambda_{pq \rightarrow q}(y_q)
\end{align}
The increase in the objective is described below using $\downarrow,\uparrow$ to denote the $\lambda$ terms before/after the update to $\lambda$.
\begin{align}
\label{improvFunEq}
(\quad \lambda^{\uparrow}_p+\sum_{\substack{q \in \mathcal{V} \\ (pq) \in \mathcal{E}}}\lambda^{\uparrow}_{pq}\quad )-(\quad \lambda^{\downarrow}_p-\sum_{\substack{q \in \mathcal{V} \\ (pq) \in \mathcal{E}}}\lambda^{\downarrow}_{pq}\quad )\\
 =(\quad \min_{y_p}\phi_p(y_p)+\sum_{\substack{ q \in \mathcal{V}  \\ (pq) \in \mathcal{E}}}\nu_{pq}(y_p) \quad )   \nonumber  \\
\nonumber-(  \min_{y_p}\phi_p(y_p) +\sum_{\substack{ q \in \mathcal{V}  \\ (pq) \in \mathcal{E}}}\lambda^{\downarrow}_{pq \rightarrow p}(y_p)+\sum_{\substack{ q \in \mathcal{V}  \\ (pq) \in \mathcal{E}}} \min_{y_p}\nu_{pq}(y_p)-\lambda^{\downarrow}_{pq \rightarrow p}(y_p) )
\end{align}
%
Observe that the iterate over $p$ in Alg \ref{MPBasic} tightens the lower bound when there is no setting of $y_p$ which minimizes $\min_{y_q}\phi_{pq}(y_p,y_q) -\lambda_{pq \rightarrow q}(y_q)$ for each $q$ and $ \min_{y_p} \phi_p(y_p)+\sum_{q, (pq) \in \mathcal{E}} \lambda_{pq \rightarrow p}(y_p)$ prior to updating $\lambda$ terms.  
\subsection{Producing an Integral Solution Given $\lambda$}
An anytime approximate solution to Eq \ref{eqn:mapGroup} is produced by independently selecting the lowest reduced cost solution $y_p$ for each $p \in \mathcal{V}$ as follows\cite{sontag}.  
%
\begin{align}
\label{eqn:EqOptInt}
y_p^* \leftarrow \min_{y_p} \phi_p(y_p) + \sum_{\substack{q \in \mathcal{V} \\ (pq) \in \mathcal{E}}} \lambda_{pq \rightarrow p}(y_p) 
\end{align}
%
%
%
%
\section{Benders Decomposition Approach}
\label{BendApproach}
In this section we study an efficient mechanism to achieve the increase in the dual objective in Eq \ref{improvFunEq}. Observe that to  compute $\nu_{pq}(y_p)$ for every possible $y_p$ one computation need be done for each member of the vector product of the state spaces of $y_p$,$y_q$ .  If we have $|\mathcal{Y}_p| = |\mathcal{Y}_q| = 15$, then we have roughly
30k states for $p$ and $q$, and enumerating the joint space of $9 \times
10^8$ states, becomes prohibitively expensive for practical applications. 
In this section we assume that all edges including $p$ are of the form $(pq)$ not $(qp)$ for notational ease.  

 To avoid enumerating the vector product of state spaces for pairs of hyper-variables we employ a  Benders decomposition \cite{benders1962partitioning} based approach.  We compute terms $\nu^-_{pq}(y_p) $ that lower bound $\nu_{pq}(y_p)$ and use $\nu^-_{pq}(y_p) $ in place of $\nu_{pq}(y_p) $ in Alg \ref{MPBasic}.  We construct $\nu^-$ so as to satisfy the following property.
\begin{align}
\label{eqValid}
 \min_{y_p}\phi_{p}(y_p)+\sum_{\substack{q \in \mathcal{V} \\ (pq) \in \mathcal{E}}}\nu^-_{pq}(y_p)= \min_{y_p}\phi_{p}(y_p)+\sum_{\substack{q \in \mathcal{V} \\ (pq) \in \mathcal{E}}} \nu_{pq}(y_p)
\end{align}
Using Eq \ref{eqValid} observe that the fixed point update using $\nu^-$ instead of $\nu$ improves the dual objective by the same amount as the standard update as describe in Eq \ref{improvFunEq}.
Since fixed point updates over $\nu^-$ increases the dual objective by exactly the same amount in a given iteration as the standard update then optimization is  guaranteed to converge to a  fixed point of dual of the local polytope relaxation. 
\subsection{Our Application of Benders Decomposition}
We now rigorously define our Benders decomposition formulation for a specific
edge $(pq) \in \mathcal{E}$ given fixed $\theta,\phi,\lambda$.  We define the set of affine functions that lower
bound $\nu_{pq}(y_p)$ as $\mathcal{Z}^{pq \rightarrow p}$
which we index by $z$ each of which is called a \textit{``Benders row"}.  We parameterize the $z$'th affine function in $\mathcal{Z}^{(pq) \rightarrow p}$ with 
$\omega_0^{z} \in \mathbb{R} $ and $\omega^{z}_{i}(x_i) \in \mathbb{R}$  for all  $i \in \mathcal{Y}_p$.

Given any  $\dot{\mathcal{Z}}^{pq \rightarrow p} \subseteq \mathcal{Z}^{pq \rightarrow p}$ we define a lower bound on $\nu_{pq}(y_p)$ denoted $\nu^-_{pq}(y_p)$ as follows.  
\begin{align}
\label{eqn:lower-envelope}
\nu_{pq}(y_p) \geq \nu^-_{pq}(y_p)= \max_{z \in \dot{\mathcal{Z}}^{pq \rightarrow p}} \omega^z_0+ \sum_{i \in \mathcal{Y}_p} \omega^z_{i} (x^{y_p}_i) 
\end{align}
%
%
%
%
To construct $\dot{\mathcal{Z}}^{pq \rightarrow p}$ for each $q$ s.t.$ (pq \in \mathcal{E})$  so as to satisfy Eq \ref{eqValid} we alternate between the following two steps.
\begin{itemize}
\item 
Select the minimizer $y^*_p$ corresponding the left hand side of Eq \ref{eqValid}.
\item Add a new Benders row to $\dot{\mathcal{Z}}^{pq \rightarrow p}$ for each neighbor $q$ that makes $\nu^-_{pq}(y^*_p)=\nu_{pq}(y^*_p)$.  As an alternative we can add a Benders row corresponding to the edge $(pq)$ in which the lower bound $\nu^-_{pq}(y^*_p)$ is loosest meaning the $q$ that maximizes $  \nu_{pq}(y^*_p)-\nu^-_{pq}(y^*_p)$.  
\end{itemize}
Termination occurs when Eq \ref{eqValid} is satisfied.  
%
%
\subsubsection{Outline of Benders Approach Section}
We outline the remainder of this section as follows.  First in Section \ref{sec:benders-gen} we produce a new Benders row corresponding to a given edge $(pq)$ that is tight at a given $y^*_p$.  Next in Section \ref{sec:solve_dual_lp_benders} we consider the use of Benders rows that are designed to speed convergence called Magnanti-Wong cuts (MWC)  or Pareto optimal cuts \cite{magnanti1981accelerating}.  Then in Section \ref{sec:reuse_rows} we show how Benders rows produced with different values of $\lambda$ than the current one can be re-used with little extra computational effort.   Finally in Section \ref{algProdNuSec} we consider  our complete algorithm for producing $\nu^-$ that satisfy Eq \ref{eqValid}. 
%
%
\subsection{Producing New Benders Rows}
\label{sec:benders-gen}
Given nascent set $\dot{\mathcal{Z}}^{pq \rightarrow p}$, and fixed $y^*_p$, we determine if the current lower bound $\nu^{-}_{pq}(y^*_p)$ is tight as follows.
\begin{align}
\label{benders-gen}
\min_{y_q} \phi_{pq}(y^*_p, y_q) - \lambda_{pq \rightarrow q}(y_q)
\end{align}
In this section we reformulate Eq \ref{benders-gen} as an ILP, then produce a tight LP relaxation with dual form that reveals a Benders row that is tight at $y_p^*$.  For ease of notation we flip $(j,i)$ for $(j,i) \in E$ where $i \in \mathcal{Y}_p,j \in \mathcal{E}$ to $(ij)$. Similarly for ease of notation we use zero valued $\theta_{ij}(x_i,x_j)$ for $(ij) \notin E$.    We use [] to denote the binary indicator function.
\begin{align}
\label{primForm}
\min_{\substack{\mu \geq 0 }} &
\sum_{y_q} -\lambda_{pq\rightarrow q}(y_q)\mu_{q}(y_q)+ \sum_{\substack{i \in \mathcal{Y}_p \\ j \in \mathcal{Y}_q \\ x_i\\x_j}} \theta_{ij}(x_i,x_j)\mu_{ij}(x_i,x_j) \\
\text{s.t.} \quad & \sum_{y_q} \mu_{q}(y_q) = 1 \nonumber \\
                  & \sum_{x_j}\mu_{ij}(x_i,x_j) = [x_i=x_i^{y^*_p}] \quad  \forall [i \in \mathcal{Y}_p,j \in \mathcal{Y}_q,x_i] \nonumber \\
                                    & \sum_{x_i}\mu_{ij}(x_i,x_j) = \sum_{y_q}[x_j=x_j^{y_q}]\mu_q(y_q) \quad \forall [i \in \mathcal{Y}_p,j \in \mathcal{Y}_q,x_j] \nonumber\\ 
& \mu_q(y_q) \in \{0,1\} \quad \forall [y_q] \nonumber \\
& \mu_{ij}(x_i,x_j) \in \{0,1\} \quad \forall [i \in \mathcal{Y}_p,j \in \mathcal{Y}_q,x_i,x_j] \nonumber
\end{align}
We assume that without loss of generality that $\theta$ is non-positive.  
In Appendix \ref{integProof} we prove that without altering the objective in Eq \ref{primForm} the integrality constraints can be forgotten and the bottom two equality constraints can be relaxed to $\sum_{x_j}\mu_{ij}(x_i,x_j) \leq  [x_i=x_i^{y^*_p}]$,$\sum_{x_i}\mu_{ij}(x_i,x_j) \leq \sum_{y_q}[x_j=x_j^{y_q}]\mu_q(y_q)$.  The corresponding optimization is below.  
\begin{align}
\mbox{Eq }\ref{primForm}=
\min_{\substack{\mu \geq 0 }} &
\sum_{y_q} -\lambda_{pq\rightarrow q}(y_q)\mu_{q}(y_q)+ \sum_{\substack{i \in \mathcal{Y}_p \\ j \in \mathcal{Y}_q \\ x_i\\x_j}} \theta_{ij}(x_i,x_j)\mu_{ij}(x_i,x_j) \\
\text{s.t.} \quad & \sum_{y_q} \mu_{q}(y_q) = 1 \nonumber \\
                  & \sum_{x_j}\mu_{ij}(x_i,x_j) \leq [x_i=x_i^{y^*_p}] \quad \forall[i \in \mathcal{Y}_p,j \in \mathcal{Y}_q,x_i] \nonumber \\
                                    & \sum_{x_i}\mu_{ij}(x_i,x_j) \leq \sum_{y_q}[x_j=x_j^{y_q}]\mu_q(y_q) \quad \forall [i \in \mathcal{Y}_p,j \in \mathcal{Y}_q,x_j] \nonumber
\end{align}
We now consider the dual problem of Eq \ref{primForm} using dual variables $\beta^0 \in \mathbb{R},\beta^1_{ij}(x_i)\in \mathbb{R}_{0+},\beta^2_{ij}(x_j) \in \mathbb{R}_{0+}$.
%
\begin{align}
\label{eqn:dualForm}
\mbox{Eq }\ref{primForm}=
\max_{\substack{\beta^0 \in \mathbb{R} \\ \beta^{1}_{ij}(x_i)\geq 0 \\ \beta^2_{ij}(x_j) \geq 0}} &
\beta^0 - \sum_{\substack{i \in \mathcal{Y}_p\\j \in \mathcal{Y}_q \\ x_i }} \beta^1_{ij}(x_i)[x_i=x^{y^*_p}_i] \\
\text{s.t.} \quad &-\lambda_{pq\rightarrow q}(y_q) - \beta^0 - \sum_{\substack{i \in \mathcal{Y}_p\\ j \in \mathcal{Y}_q \\x_j }}  [x_j=x_j^{y_q}]\beta^2_{ij}(x_j) \geq 0 \quad \forall y_q \nonumber \\
                  & \theta_{ij}(x_i,x_j) +\beta^1_{ij}(x_i)+\beta^2_{ij}(x_j)\geq 0 \quad \forall[ i \in \mathcal{Y}_p, j \in \mathcal{Y}_q,x_i,x_j] \nonumber
\end{align} 
Observe that given fixed $\beta$ the objective in Eq \ref{eqn:dualForm}  is an affine function of $x_i\quad \forall  i \in \mathcal{Y}_p$.  
Thus when dual variables are optimal Eq \ref{eqn:dualForm} represents a new
Benders row that we add to $\dot{\mathcal{Z}}^{pq \rightarrow p}$ that makes the lower bound in Eq \ref{eqn:lower-envelope} tight at $y_p^*$.  Let us
denote the new Benders row as $z^*$ which we construct from $\beta$ as follows.
\begin{align}
\label{eqn:omega0-from-deltas}
\omega_0^{z^*} &= \beta^0 \\
\nonumber \omega_i^{z^*}(x_i) &= -\sum_{\substack{j \in \mathcal{Y}_q}}\beta^1_{ij}(x_i) \quad \forall [i \in \mathcal{Y}_p,x_i]
\end{align}
\subsection{Magnanti-Wong Cuts}
\label{sec:solve_dual_lp_benders}
One can directly solve Eq \ref{eqn:dualForm} via an
off-the-shelf LP solver, which gives a tight lower bound for a given $y^*_p$.  However, ideally we want this new Benders row to also give a
good (not terribly loose) lower bound for other selections of $y_p$, so that we can use as few computations as
possible to satisfy Eq \ref{eqValid}. 

Approaches for generating Benders rows that produce good lower bounds are called Pareto optimal cuts or Magnanti-Wong cuts
\cite{magnanti1981accelerating} (MWC) in the operations research literature.  We generate a MWC by adding regularization with tiny positive weight
$\epsilon$ to prefer smaller values of  $\beta^1$ as follows.
\begin{align}
\label{paretoQpLpForm}
- \epsilon \sum_{i \in \mathcal{Y}_p}\min_{x_i}\sum_{\substack{j \in \mathcal{Y}_q}}\beta^1_{ij}(x_i)
\end{align}
\subsubsection{Specialization for Binary Problems}
In Appendix  \ref{MagWongBin} derive an efficient exact procedure for  producing MWC for binary problems.  We write the findings below.  
Recall that WLOG at exactly one entry of $\theta_{ij}(x_i,x_j)$ is non-zero and that entry is non-positive (See Appendix \ref{binarprop} for details).  We denote the indexes corresponding to that entry as $x^{ij}_i$,$x^{ij}_j$ respectively.  We write fast computation of $\beta$ below using $\beta^3_j(x_j)=\sum_{i \in \mathcal{Y}_p}\beta^1_{ij}(x^{ij}_i)[x_j=x^{ij}_j]$ as follows using helper constants $Q_j(x_j),h_{ij}$.  
\begin{align}
\label{magWongBinEq}
\beta_0+\epsilon \max_{\substack{Q_j(x_j)\geq \beta^{3}_{j}(x_j) \geq 0}} &
-\sum_{\substack{j \in \mathcal{Y}_q \\ x_j}}\beta^3_{j}(x_j)\\
\text{s.t.} \quad & -\lambda_{pq\rightarrow q}(y_q) - \beta^0 + \sum_{\substack{i \in \mathcal{Y}_p\\j \in \mathcal{Y}_q }} [x^{ij}_j=x_j^{y_q}](\theta_{ij}(x_i^{ij},x_j^{ij})+\beta^3_{j}(x_j^{ij})h_{ij}) \geq 0 \quad \forall y_q \nonumber 
\end{align} 
where \footnote{Examine the mapping  $\beta^3_j(x_j)=\sum_{i \in \mathcal{Y}_p}\beta^1_{ij}(x^{ij}_i)[x_j=x^{ij}_j]$ and observe that when $Q_j(x_j)=0$ then $\beta^3(x_j)=0$ not undefined.  Similarly in the case that $Q_j(x_j)=0$ then $\beta^3_{j}(x_j)h_{ij}=0$ not undefined.}:
\begin{align}
 \nonumber \beta^3_j(x_j)=\sum_{i \in \mathcal{Y}_p}\beta^1_{ij}(x^{ij}_i)[x_j=x^{ij}_j] \quad \forall [j\in \mathcal{Y}_q,x_j]\\
\nonumber \beta^1_{ij}(x^{ij}_i)=\beta^3_{j}(x^{ij}_j)h_{ij} \quad \forall [i\in \mathcal{Y}_p,j \in \mathcal{Y}_q]\\
\nonumber \beta^2_{ij}(x^{ij}_j)=-\theta_{ij}(x^{ij}_i,x^{ij}_j)-\beta^1_{ij}(x^{ij}_i) \quad  [i\in \mathcal{Y}_p,j \in \mathcal{Y}_q]\\
\nonumber h_{ij}=\frac{-\theta_{ij}(x^{ij}_i,x^{ij}_j) [x^{ij}_i \neq x^{y^*_p}_i]  }{Q_j(x^{ij}_j)} \quad  [i\in \mathcal{Y}_p,j \in \mathcal{Y}_q]\\
\nonumber Q_j(x_j)=\sum_{i \in \mathcal{Y}_p}-\theta_{ij}(x^{ij}_i,x^{ij}_j) [ x^{ij}_i \neq x^{y^*_p}_i] [ x^{ij}_j = x_j]  \quad  \forall [j\in \mathcal{Y}_q,x_j]\\
\nonumber \beta_0= \min_{y_q} \phi_{pq}(y^*_p, y_q) - \lambda_{pq \rightarrow q}(y_q)
\\
\nonumber \beta^1_{ij}(x_i)=0 \quad \forall x_i \mbox{  s.t. } x_i \neq x^{ij}_i \\
\nonumber \beta^2_{ij}(x_j)=0 \quad \forall x_j \mbox{  s.t. } x_j \neq x^{ij}_j 
\end{align}
We often observe in a pre-solve that setting $\beta^3$ to the zero vector satisfies a large portion of the constraints over $y_q$.  Therefor these constraints are satisfied for all $\beta^3$ and hence we remove them from consideration. We may choose to use a mixture of the L1 and L2 norm over optimization of $\beta^3$ parameterized by $\alpha \in [0,1]$ as discussed in the Appendix \ref{MagWongBin}.  We write the corresponding objective below.  
\begin{align}
\beta_0+\epsilon \max_{\substack{Q_j(x_j)\geq \beta^{3}_{j}(x_j) \geq 0}} &
-\sum_{\substack{j \in \mathcal{Y}_q \\ x_j}}\alpha\beta^3_{j}(x_j)+(1-\alpha)\beta^3_{j}(x_j)\beta^3_{j}(x_j)
\end{align}
%
%
%
%
%
%
\subsubsection{Reverse Magnanti-Wong Cuts}%
In this section we propose a cut that avoids solving the LP in Eq \ref{magWongBinEq} or Eq \ref{eqn:dualForm}.  These correspond to the least binding cut that is tight at $y^*_p$  and we refer to them as Reverse Magnanti-Wong Cuts (RMC).  The RMC corresponds to setting $ \beta^3_{j}(x_j) \leftarrow    Q_j(x_j) $ for all $[j\in \mathcal{Y}_q,x_j]$ which results in a feasible solution to Eq \ref{magWongBinEq}.  
%
RMC can be used for non-binary problems by setting $\beta$ as follows (see Appendix \ref{revDeriv} for details):
\begin{itemize}
\item
$\beta^0\leftarrow \min_{y_q} \phi_{pq}(y^*_p, y_q) - \lambda_{pq \rightarrow q}(y_q)$.
\item
 $\beta^1_{ij}(x_i)\leftarrow -[x_i \neq x_i^{y_p^*}]\min_{x_j}\theta_{ij}(x_i,x_j)$ \quad $\forall[i\in \mathcal{Y}_p,j\in \mathcal{Y}_q,x_i]$.  
\item 
 $\beta^2_{ij}(x_j)\leftarrow -\min_{x_i}\theta_{ij}(x_i,x_j)+\beta^1_{ij}(x_i)$ \quad  $\forall[i\in \mathcal{Y}_p,j\in \mathcal{Y}_q,x_j]$.  
\end{itemize}
\subsection{Recycling Benders Rows After Updates to $\lambda$}
\label{sec:reuse_rows}
In Alg \ref{MPBasic}, updates to $\lambda_{pq \rightarrow p}(y_p)$ terms; require
re-computing all $\nu$ terms from scratch.  In this section we save computation time by re-using Benders rows from the previous iterations before generating new Benders rows. 

Updates to $\lambda_{pq \rightarrow q}(y_q)$ may make the Benders rows in
$\dot{\mathcal{Z}}^{pq\rightarrow p}$ generated in previous iterations no longer lower bound $\nu_{pq}(y_p)$.  Instead of constructing $\dot{\mathcal{Z}}^{pq \rightarrow p}$ from scratch every iteration, we re-use $\beta$ terms produced by previous iterations.  Specifically for each Benders row in $\dot{\mathcal{Z}}^{pq\rightarrow p}$
we leave $\beta^1,\beta^2$ unchanged and set the $\beta^0$ term to the maximum feasible value.  We write the corresponding update below.  
\begin{align}
\label{update_delta_0}
\beta^0 \leftarrow \min_{y_q}-\lambda_{pq \rightarrow q}(y_q)  - \sum_{\substack{i \in \mathcal{Y}_p \\ j \in \mathcal{Y}_q  \\x_j }}  [x_j=x_j^{y_q}]\beta^2_{ij}(x_j) 
\end{align}
Solving Eq \ref{update_delta_0} is  accelerated by storing $\sum_{\substack{ij,x_j }}  [x_j=x_j^{y_q}]\beta^2_{ij}(x_j) $ for each selection of $y_q$.
In practice we observe that re-using Benders rows gives vast speed-ups compared
with constructing Benders rows from scratch for each iteration of
Alg \ref{MPBasic}.
\subsubsection{Recycling Reverse Magnanti-Wong Cuts }
The RMC have the following useful property in addition to being fast to compute.  Notably they remain tight after changes in $\lambda$ at the $y_p^*$ that was used to generate them given that the corresponding $\beta^0$ term is first updated as described in Eq \ref{update_delta_0}.  This is because the $\beta^{1},\beta^{2},\beta^{3}$ do not vary with $\lambda$.
%
\subsubsection{Minor point:  Sharing elements between $\dot{\mathcal{Z}}^{pq\rightarrow p}$, and $\dot{\mathcal{Z}}^{pq\rightarrow q}$}
To provide additional speed we share all elements between $\dot{\mathcal{Z}}^{pq \rightarrow p}$ and $\dot{\mathcal{Z}}^{pq \rightarrow q}$. For a given pair of elements $z_p,z_q$ that correspond their $\beta^1,\beta^2$ terms are identical however the $\beta^{0}$ terms differ as determined by Eq \ref{update_delta_0}.  
\subsection{Algorithm for computing $\nu^-$ }
\label{algProdNuSec}

\begin{figure}
\begin{algorithm}[H]
 \caption{Benders Formulation for Passing Message on Node $p$}
\begin{algorithmic}[1]
\For{$q \quad \mbox{ s.t. }\quad   (pq)  \in\mathcal{E}$}\label{startPre}
\For{$z \in \dot{\mathcal{Z}}^{pq \rightarrow p}$}\label{startZpre}
\State $\beta^{0z} \leftarrow \min_{y_q}-\lambda_{pq \rightarrow q}(y_q)  - \sum_{\substack{ij,x_j }}  [x_j=x_j^{y_q}]\beta^{2z}_{ij}(x_j) $
\State $\omega^{z}_0 \leftarrow \beta^{0z}$
\EndFor \label{doneZpre}
\For{$y_p$}\label{pUpdatePre}
\State $\nu^-_{pq}(y_p)\leftarrow \max_{z \in \dot{\mathcal{Z}}^{pq \rightarrow p}}\omega^{z}_0+\sum_{i \in \mathcal{Y}_p}\omega_i^z(x^{y_p}_i)$
\EndFor \label{donepUpdatePre}
\EndFor \label{donePre}
\Repeat \label{startOpt}
\State $y^*_p\leftarrow \mbox{arg}\min_{y_p}\phi_p(y_p)+\sum_{q; (pq) \in \mathcal{E}}\nu^-_{pq}(y_p)$\label{computeyp}
\State $q^* \leftarrow \mbox{arg}\max_{\substack{q \\ (pq) \in \mathcal{E}}}(\min_{y_q} \phi_{pq}(y_p^*,y_q)-\lambda_{pq \rightarrow q}(y_q))- \nu^-_{pq}(y_p^*)$ \label{computeyq}
\State $z^* \leftarrow$ Solve Eq \ref{eqn:dualForm} on edge $(pq^*)$ given $y_p^*$ \label{startAddz}
\State $\dot{\mathcal{Z}}^{pq^* \rightarrow p} \leftarrow \dot{\mathcal{Z}}^{pq^* \rightarrow p} \cup z^*$\label{doneAddz}
\For{$y_p$}\label{startUpdateMM}
\State $\nu^-_{pq}(y_p)\leftarrow \max(\nu^-_{pq}(y_p); \quad \omega^{z^*}_0+\sum_{i \in \mathcal{Y}_p}\omega_i^{z^*}(x^{y_p}_i))$
\EndFor \label{endUpdateMM}
 \Until{ $\min_{y_p}\phi_{p}(y_p)+\sum_{q;(pq) \in \mathcal{E}}\nu^-_{pq}(y_p)= \phi_{p}(y^*_p)+\sum_{q;(pq) \in \mathcal{E}}\nu_{pq}(y^*_p)$ }\label{endOpt} 
\end{algorithmic}
\label{MPBender}
  \end{algorithm}
  \caption{We display our algorithm for computing $\nu^-$.  We use $\beta^{0z},\beta^{1z},\beta^{2z}$ to describe the $\beta$ terms associated with Benders row $z$.}
\end{figure}
Our algorithm proceeds by iteratively testing if Eq \ref{eqValid} holds and generating Benders rows corresponding to a minimizer $y_p$ of $\phi_{p}(y_p)+\sum_{\substack{q \\ (pq) \in \mathcal{E}}}\nu^-_{pq}(y_p)$.  Before we generate new Benders rows we re-cycle the ones already generated as described in Eq \ref{update_delta_0}.  We display our algorithm below in Alg \ref{MPBender} with line by line explanation below.  
\begin{itemize}
\item \textbf{ \ref{startPre}-\ref{donePre}} Iterate over neighbors of $p$:  Update the Benders rows associated with $\dot{\mathcal{Z}}^{pq\rightarrow p}$.  Update the message terms $\nu^-$.
%
\begin{itemize}
\item \textbf{\ref{startZpre}-\ref{doneZpre}:  }Iterate over $z \in \dot{\mathcal{Z}}^{pq\rightarrow p}$: Compute intercept of Benders row $z$ following Eq \ref{update_delta_0}.  
\item \textbf{\ref{pUpdatePre}-\ref{donepUpdatePre}:  }  Update  $\nu^-_{pq}(y_p)$ for each $y_p$ using Eq \ref{eqn:lower-envelope}. 
\end{itemize}
\item \textbf{\ref{startOpt}-\ref{endOpt}:  }  Alternate between selecting $y^*_p$ that minimizes the left hand side of Eq $\ref{eqValid}$ and adding new Benders rows until Eq \ref{eqValid} is satisfied.
\begin{itemize}
\item \textbf{\ref{computeyp}:  } Select $y_p^*$ that minimizes left hand side of Eq $\ref{eqValid}$.
\item \textbf{\ref{computeyq}:  }  Select the edge $(pq)$ associated with the largest difference $\nu_{pq}(y^*_p)-\nu^-_{pq}(y^*_p)$.  
\item \textbf{\ref{startAddz}-\ref{doneAddz}:  }  Compute new Benders row  $z^*$ according to Eq \ref{eqn:dualForm} with choice of MWC or RMC. 
\item \textbf{\ref{startUpdateMM}-\ref{endUpdateMM}:  } Iterate over $y_p$ and update $\nu^-_{pq}(y_p)$ using new Benders row $z^*$.
\end{itemize}
\end{itemize}
\subsubsection{Initializing $\dot{\mathcal{Z}}^{pq \rightarrow p}$}
At the beginning of Alg \ref{MPBasic}  for each $(pq) \in \mathcal{E}$ we initialize $\dot{\mathcal{Z}}^{pq \rightarrow p}$ with the RMC corresponding to $y^*_p= \min_{y_p}\phi_p(y_p)$ which is a guess at the selection of $y_p$ associated with the optimizer of Eq \ref{eqn:mapGroup}. 
\section{Experiments}
\label{exper}
In this section we consider the performance of our approach on binary MRFs corresponding to the work on multi-person pose estimation of \cite{wang2017efficient}.  We outline this experiments section as follows.  In Section \ref{domainProb} we describe our problem instances and their origin in human pose estimation.  In Section \ref{conHyper} we describe our construction of hyper-variables.  In Section \ref{variantSec} we describe the variants of our approach that we compare.  In Section \ref{experPerf} we describe experiments performed and show plots corresponding to timing and cost.  We compare against a baseline of doing standard MPLP\cite{sontag,sontag2011introduction}.  Finally we make observations about the relative performance of the variants of our approach and the baseline.  All comparisons are done using MATLAB with a shared code (across variants and baseline) implementing Alg \ref{MPBasic}.
\subsection{Problem Domain}
\label{domainProb}
We compute the lowest cost human pose in each of the problem instances which corresponds to computing the MAP of a binary MRF (see \cite{wang2017efficient} for details on problem instances).  
In each problem instance there is a surjection of variables to human body parts:  head, neck and the left/right of  wrist, elbow, shoulder, hip, knee, ankle.  Binary variables being set to 0/1 indicate their exclusion/inclusion in the human pose.   The human body is modeled as a conditional tree structure with the neck connected to every body part but the common tree structure followed otherwise. 

In \cite{wang2017efficient} the authors simplify the MAP problem by conditioning on the variables corresponding to the neck then solving the MAP problem as a dynamic program sped up using a Nested benders decomposition\cite{birge1985decomposition}.  In this case each body part defines a hyper-variable in the dynamic program with state space limited to 50,000 and the neck is further limited (see \cite{wang2017efficient} for details).  In our experiments we dispense with these limitations and do not condition on the neck.
\subsection{Constructing Hyper-Variables}
\label{conHyper}
Construction of hyper-variables is parameterized by positive user defined integer $M$ which defines the maximum number of variables associated with a hyper-variable.  Since the local polytope relaxation is tight on a tree we design $\mathcal{E}$ to be as tree like as possible.  To this end each hyper-variable consists of exactly one type of body part.  For a given body part associated with $b$ variables then there are $\lfloor \frac{b}{M}\rfloor$ hyper-variables with $M$ members and one hyper-variable with the remaining members.  The variables corresponding to a given body part are assigned arbitrarily to the corresponding hyper-variables given the size constraints. 
%
%
\subsection{Variants of our Approach}
\label{variantSec}
We consider variants of our approach in Alg \ref{MPBender} parameterized by two terms.  
\begin{itemize}
\item Whether to generate MWC corresponding to each $(pq)$ pair or only over the most violated term as is done in Alg \ref{MPBender}.
\item Which setting of $\alpha$ should be used when generating a MWC.  For a given edge multiple such Benders rows may be generated.  
\end{itemize}
In our experiments we consider ten variants which are described in Table \ref{variantTable}.  
\begin{figure}
\begin{tabular}{l*{7}{c}r}
Algorithm              & Color & Symbol & (solid=0 or dotted=1) & $\alpha=1$ & $\alpha=0$  & $\alpha=\frac{1}{2}$ & RMC \\
\hline
1  & red          & o & 0 & 1 & 0 & 0 & 0  \\
2  & green      & o & 0 & 0 & 1 & 0 & 0  \\
3  & blue        & o & 0 & 0 & 0 & 1 & 0  \\
4  & magenta & o & 0 & 0 & 0 & 0 & 0  \\
5  & cyan       & o & 0 & 1 & 1 & 1 & 0  \\
6   & red          & x & 1 & 1 & 0 & 0 & 1  \\
7   & green      & x & 1 & 0 & 1 & 0 & 1  \\
8   & blue        & x & 1 & 0 & 0 & 1 & 1  \\
9   & magenta & x & 1 & 0 & 0 & 0 & 1  \\
10 & cyan       & x & 1 & 1 & 1 & 1 & 1  
\end{tabular}
\caption{Each row of the table above describes a variant.  The first four columns associate  color, symbol, and line style used for plotting comparisons.  The columns $\alpha=1,0,\frac{1}{2}$ indicates whether the given $\alpha$ is used to generate rows associated with the edge $(pq^*)$.  We use RMC =1 to indicate that reverse Magnanti-Wong cuts are generated on all edges $(pq)$  and set RMC=0 if reverse Magnanti-Wong cuts are generated on $pq^*$ only.}
\label{variantTable}
\end{figure}

\begin{figure}
\begin{tabular}{l*{5}{c}r}
Variant/Percentile;M=5    &80&90&93&95&98&100\\
\hline
1&0.000634&0.0174&0.0434&0.0531&0.125&1.95\\
2&3.81e-05&0.0117&0.0181&0.0517&0.126&1.55\\
3&4.09e-05&0.0117&0.0169&0.0631&0.204&1.55\\
4&0.000634&0.0174&0.0434&0.0531&0.125&1.95\\
5&3.81e-05&0.0117&0.0169&0.0631&0.204&1.55\\
6&1.37e-06&0.000332&0.00899&0.038&0.124&2.02\\
7&5.96e-07&8.29e-05&0.00925&0.034&0.179&2.42\\
8&6.12e-07&8.29e-05&0.00925&0.034&0.179&2.42\\
9&1.37e-06&0.000332&0.00899&0.038&0.124&2.02\\
10&5.96e-07&8.29e-05&0.00925&0.034&0.179&2.42\\
11&3.45e-16&6.13e-16&0.00807&0.0361&0.111&0.164
\end{tabular}
\begin{tabular}{l*{5}{c}r}
Variant/Percentile;M=16    &80&90&93&95&98&100\\
\hline
1&1.7e-16&3.25e-16&6.98e-07&0.00149&0.0237&0.0838\\
2&1.08e-14&2e-10&8.57e-07&2.06e-06&0.0343&0.119\\
3&4.61e-16&4.81e-13&6.4e-07&2.06e-06&0.0343&0.119\\
4&1.7e-16&3.25e-16&6.98e-07&0.00149&0.0237&0.0838\\
5&3.35e-16&3.93e-13&6.4e-07&2.06e-06&0.0343&0.119\\
6&1.33e-16&2.16e-16&3.19e-16&2.1e-07&0.015&0.0828\\
7&1.97e-16&5.42e-13&1.65e-06&1.57e-05&0.0165&0.0836\\
8&2.12e-16&1.54e-14&1.65e-06&1.57e-05&0.0165&0.0836\\
9&1.33e-16&2.16e-16&3.19e-16&2.1e-07&0.015&0.0828\\
10&1.94e-16&2.69e-15&1.65e-06&1.57e-05&0.0165&0.0836\\
11&1.95e-16&3.25e-16&4.02e-16&4.45e-16&0.0145&0.0828
\end{tabular}
\caption{Percentiles of Normalized Gap across Algorithm for M=5,16.}
\label{energyTable}
\end{figure}
%
%
%
%
\begin{figure}
\includegraphics[width=.8\textwidth,trim={1.5cm 7cm 1cm 6cm},clip]{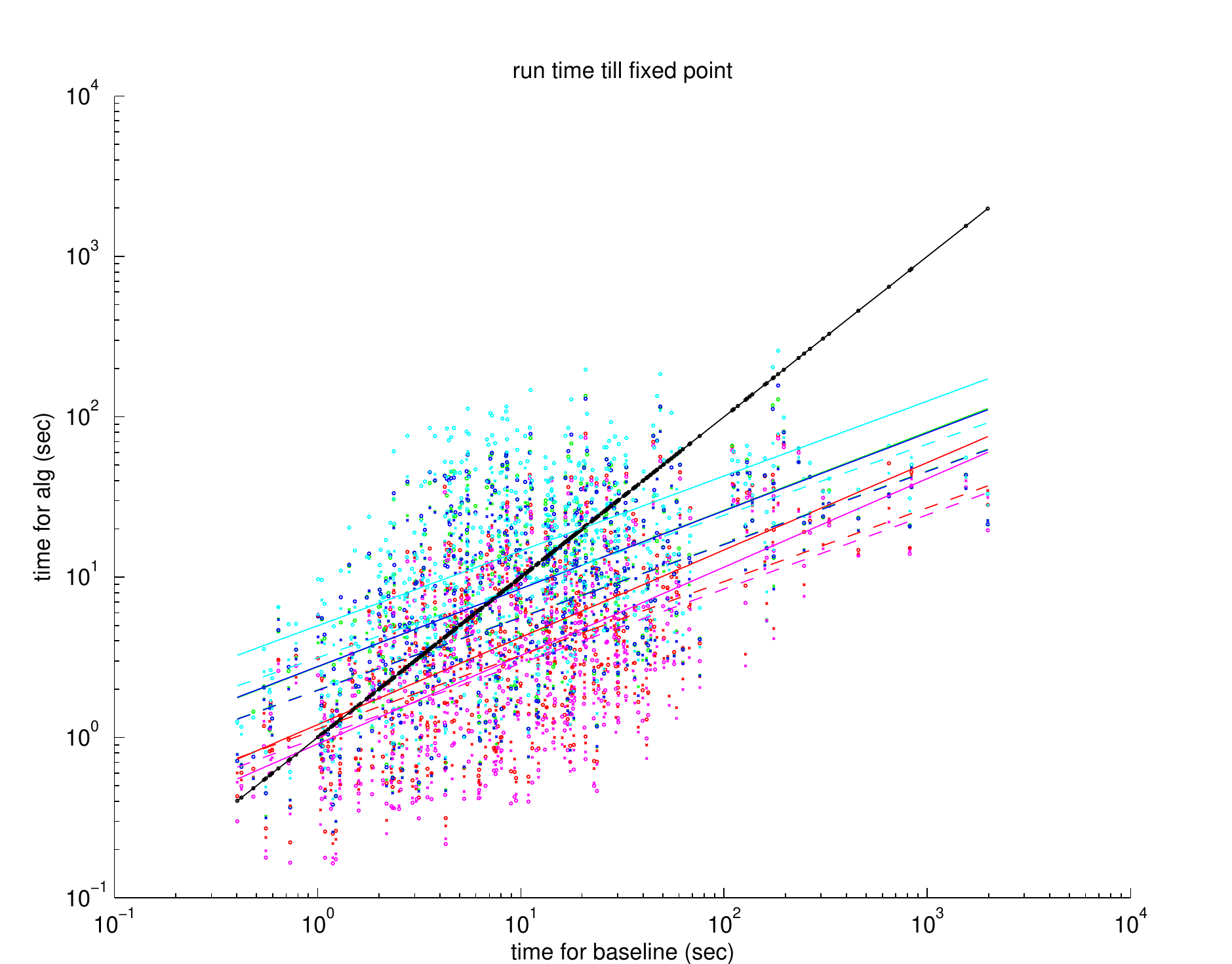}
\includegraphics[width=.8\textwidth,trim={1.5cm 7cm 1cm 6cm},clip]{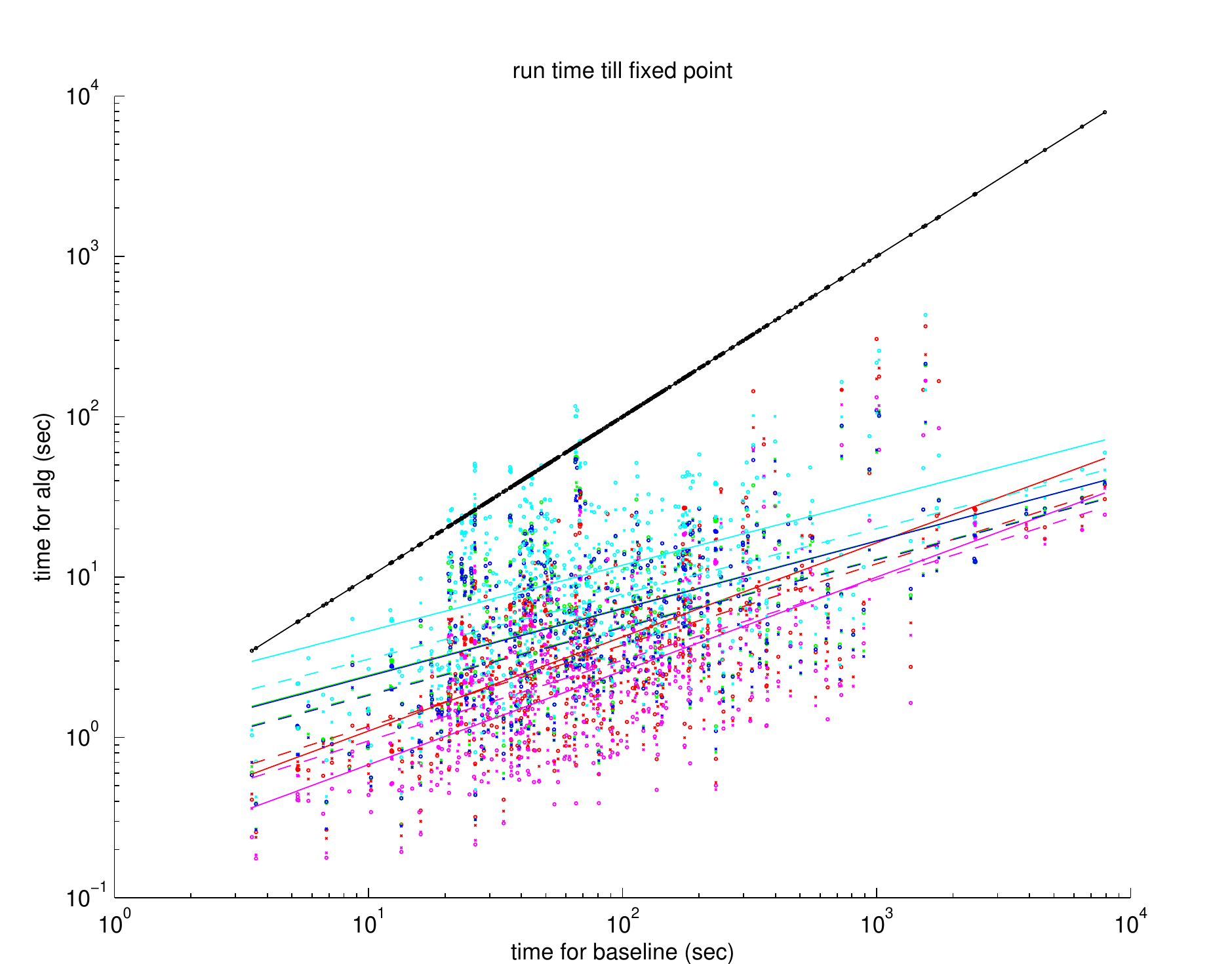}
     \caption{Timing comparison of variants relative to performance of baseline  MPLP.   Each data point is defined by a variant, time for the baseline (x axis),time for that variant (y axis).  Each plotted line is the line of best fit across calls to the given variant (or baseline).  Color/symbol legend is in Table \ref{variantTable} .  \textbf{Top:}  Shows plot over M=[5,7,9,11].  \textbf{Bottom:}  Shows plot over M=[13,14,15,16].  We observe that the green lines/points are almost occluded by the corresponding blue ones indicating highly similar performance.  }
\label{timingEngComp2}
\end{figure}
\subsection{Experiments Performed}
\label{experPerf}
We consider a random subset of ninety seven problem instances from \cite{wang2017efficient}.  We capped the number of message passing iterations (outer loop steps in Alg \ref{MPBasic}) to 400 but terminate when no progress is made in a given iteration of message passing.  We consider M=[5,7,9,11,13,14,15,16].  We (scatter) plot the total time consumed relative to the MPLP baseline across problem instances in Figure \ref{timingEngComp2}.  

We consider the difference between the upper and lower bound at convergence across variant.  We describe this in terms of a normalized gap which is the difference between the upper and lower bounds divided by the absolute value of the lower bound.   We use the lowest cost upper bound computed across iterations of the outer loop for a given variant (or MPLP).  In Table \ref{energyTable} show display for normalized gap for various percentiles achieved for M=[5,16].  
\subsection{Observations}
\label{observeMade}
We now summarize the findings of our experiments.  

  With regards to the difference between upper/lower bounds as a function of variant we observe little difference between the variants and the baseline.  

With regards to timing we observe that the performance of the variants differ wildly on any given instance.  However relative to the baseline we observe the following common result.  The longer the baseline takes larger the relative speedup introduced across our variants. However all of our variants tend to outperform the baseline on instances where the baseline takes substantial time.

In our experiments we observe that the variants employing solely RMC lead to the best performance.  This contrasts heavily with the work of \cite{wang2017efficient} which only makes use of Benders rows corresponding to L1 regularization ($\alpha=1$).   
%
%
\section{Conclusion}
\label{conc}
In this paper we introduce a mechanism for inference in MRFs with low state space variables under the message passing framework.   We first group the variables into hyper-variables and apply message passing over the hyper-variables.  In this domain it is not feasible to compute min-marginals exactly motivating the use of Benders decomposition. We use Benders decomposition to lower bound the min-marginals by an upper envelope of affine functions of the variables composing the hyper-variable receiving the message.

These envelopes are then used in place of the min-marginals in message passing.  We guarantee convergence to a fixed point of the dual problem and observe that our approach greatly accelerates inference relative to standard message passing.  We recycle Benders rows after updates to $\lambda$ and generate Magnanti-Wong cuts using an efficient mechanism targeted to binary problems.

In future work we intend to tailor the relaxation to be as tight as possible by constructing $\mathcal{Y}$ as inference progresses.  A trivial approach is to merge hyper-variables as needed to tighten the relaxation.  Further challenges manifest in extending our framework to permit overlapping sets in $\mathcal{Y}$.
\bibliographystyle{ieee}
\bibliography{col_gen_bib}
\appendix
\section{Integrality Proof}
\label{integProof}
 In this section we prove that without altering the objective in Eq \ref{primForm} the integrality constraints can be forgotten and the second two equality constraints can be relaxed to $\sum_{x_j}\mu_{ij}(x_i,x_j) \leq  [x_i=x_i^{y_p}]$,$\sum_{x_i}\mu_{ij}(x_i,x_j) \leq \sum_{y_q}[x_j=x_j^{y_q}]\mu_q(y_q)$.  

 First observe in Eq \ref{primForm} terms corresponding to  $[x_i\neq x_i^{y^*_p}]$ can be ignored since they are known to be zero.  We now rewrite optimization.  
\begin{align}
\label{intOpPor}
\min_{\substack{\mu \geq 0 }} &
\sum_{y_q} -\lambda_{pq\rightarrow q}(y_q)\mu_{q}(y_q)+ \sum_{\substack{i \in \mathcal{Y}_p \\ j \in \mathcal{Y}_q \\x_j}} \theta_{ij}(x_i^{y^*_p},x_j)\mu_{ij}(x^{y_p^*}_i,x_j) \\
\text{s.t.} \quad & \sum_{y_q} \mu_{q}(y_q) = 1 \nonumber \\
                  & \sum_{x_j}\mu_{ij}(x^{y^*_p}_i,x_j) = 1, \quad [i \in \mathcal{Y}_p,j \in \mathcal{Y}_q] \nonumber \\
                                    & \mu_{ij}(x^{y^*_p}_i,x_j) = \sum_{y_q}[x_j=x_j^{y_q}]\mu_q(y_q), \quad \forall [i \in \mathcal{Y}_p,j \in \mathcal{Y}_q,x_j] \nonumber\\ 
& \mu_q(y_q) \in \{0,1\} \quad \forall [y_q] \nonumber \\
& \mu_{ij}(x_i,x_j) \in \{0,1\} \quad \forall [i \in \mathcal{Y}_p,j \in \mathcal{Y}_q,x_i,x_j] \nonumber
\end{align}
%
We outline the remainder of this  section into two parts.  In Section \ref{intOpPor} we establish that the integrality constraints can be removed from Eq \ref{intOpPor}.  Then in Section \ref{relaxPor} we relax the second two equality constraints in Eq \ref{primForm} to inequalities.  
\subsection{Relaxing Integrality}
\label{intOpPor}
We  substitute $\mu_{ij}(x^{y^*_p}_i,x_j)$ with $ \sum_{y_q}[x_j=x_j^{y_q}]\mu_q(y_q)$ producing the following optimization.   
 \begin{align}
\min_{\substack{\mu \geq 0 }} &
\sum_{y_q} -\lambda_{pq\rightarrow q}(y_q)\mu_{q}(y_q)+ \sum_{\substack{i \in \mathcal{Y}_p \\ j \in \mathcal{Y}_q \\x_j}} \theta_{ij}(x_i^{y_p^*},x_j)(\sum_{y_q}[x_j=x_j^{y_q}]\mu_q(y_q))\\
\text{s.t.} \quad & \sum_{y_q} \mu_{q}(y_q) = 1 \nonumber \\
                  & \sum_{x_j}\sum_{y_q}[x_j=x_j^{y_q}]\mu_q(y_q)= 1, \quad [i \in \mathcal{Y}_p,j \in \mathcal{Y}_q] \nonumber \\
& \mu_q(y_q) \in \{0,1\} \quad \forall [y_q] \nonumber 
\end{align} 
 Observe that $\sum_{x_j}\sum_{y_q}[x_j=x_j^{y_q}]\mu_q(y_q)= \sum_{y_q} \mu_{q}(y_q) $.  Thus we eliminate the redundant constraint $\sum_{x_j}\sum_{y_q}[x_j=x_j^{y_q}]\mu_q(y_q)=1$ producing the following LP. 
  \begin{align}
   \label{relInPorM}
\min_{\substack{\mu \geq 0 }} &
\sum_{y_q} -\lambda_{pq\rightarrow q}(y_q)\mu_{q}(y_q)+ \sum_{\substack{i \in \mathcal{Y}_p \\ j \in \mathcal{Y}_q \\x_j}} \theta_{ij}(x_i^{y_p^*},x_j)(\sum_{y_q}[x_j=x_j^{y_q}]\mu_q(y_q))\\
\text{s.t.} \quad & \sum_{y_q} \mu_{q}(y_q) = 1 \nonumber \\
& \mu_q(y_q) \in \{0,1\} \quad \forall [y_q] \nonumber 
\end{align}
Observe that Eq \ref{relInPorM} is totally unimodular via the consecutive ones property so we can relax integrality explicitly without loosening the optimization in Eq \ref{primForm}.
 \subsection{Relaxing Equality}
\label{relaxPor}
 We begin by rewriting Eq \ref{primForm} with integrality relaxed.  
  \begin{align}
\min_{\substack{\mu \geq 0 }} &
\sum_{y_q} -\lambda_{pq\rightarrow q}(y_q)\mu_{q}(y_q)+ \sum_{\substack{i \in \mathcal{Y}_p \\ j \in \mathcal{Y}_q \\x_j}} \theta_{ij}(x_i^{y_p^*},x_j)\mu_{ij}(x_i^{y_p^*},x_j) \\
\text{s.t.} \quad & \sum_{y_q} \mu_{q}(y_q) = 1 \nonumber \\
                  & \sum_{x_j}\mu_{ij}(x^{y^*_p}_i,x_j) = 1 \quad \forall [i \in \mathcal{Y}_p,j \in \mathcal{Y}_q] \nonumber \\
                                    & \mu_{ij}(x^{y^*_p}_i,x_j) = \sum_{y_q}[x_j=x_j^{y_q}]\mu_q(y_q), \quad \forall [i \in \mathcal{Y}_p,j \in \mathcal{Y}_q,x_j] \nonumber
\end{align}
 In Section \ref{intOpPor} we eliminated the constraint $\sum_{x_j}\mu_{ij}(x^{y^*_p}_i,x_j) = 1$ since it is shown to be redundant.  Hence we can relax it to $\sum_{x_j}\mu_{ij}(x^{y^*_p}_i,x_j) \leq1$ without loosening  the relaxation.  We write the corresponding optimization below ignoring $\sum_{x_j}\mu_{ij}(x^{y^*_p}_i,x_j) = 1$.  
  \begin{align}
\min_{\substack{\mu \geq 0 }} &
\sum_{y_q} -\lambda_{pq\rightarrow q}(y_q)\mu_{q}(y_q)+ \sum_{\substack{i \in \mathcal{Y}_p \\ j \in \mathcal{Y}_q \\x_j}} \theta_{ij}(x_i^{y_p^*},x_j)\mu_{ij}(x_i^{y_p^*},x_j) \\
\text{s.t.} \quad & \sum_{y_q} \mu_{q}(y_q) = 1 \nonumber \\
 & \mu_{ij}(x^{y^*_p}_i,x_j) = \sum_{y_q}[x_j=x_j^{y_q}]\mu_q(y_q), \quad \forall [i \in \mathcal{Y}_p,j \in \mathcal{Y}_q,x_j] \nonumber
\end{align}
Since $\mu_{ij}(x^{y^*_p}_i,x_j)$ for each $j,x_j$  are each bound independently given $\mu_q$ and non-positive in the objective then it is optimal to set $ \mu_{ij}(x^{y^*_p}_i,x_j) = \sum_{y_q}[x_j=x_j^{y_q}]\mu_q(y_q)$ even if equality constraint is relaxed to $ \mu_{ij}(x^{y^*_p}_i,x_j) \leq \sum_{y_q}[x_j=x_j^{y_q}]\mu_q(y_q)$.
%
%
 %

\section{Efficient Computation of Magnanti-Wong Cuts for Binary Problems}
\label{MagWongBin}
In this section we consider an efficient mechanism to generate Magnanti-Wong cuts for binary problems.  We use the derivation of \cite{wang2017efficient} Appendix B with notation from our document.  

Recall that only one entry of $\theta_{ij}(x_i,x_j)$ is non-zero.  We denote the corresponding values as $x^{ij}_i$,$x^{ij}_j$ respectively.  
Note that in Eq \ref{eqn:dualForm} that the constraint $\theta_{ij}(x_i,x_j)+\beta^1_{ij}(x_i)+\beta^2_{ij}(x_j) \geq 0$ for $x_i\neq x^{ij}_i$ or $x_j \neq x^{ij}_j$  is always satisfied since $\beta^1,\beta^2$ are non-negative.  

We now simplify our notation.  We use $\theta_{ij} $ to denote $ \theta_{ij}(x^{ij}_i,x^{ij}_j)$.  Similarly we  use $\beta^1_{ij},\beta^2_{ij}$  to denote the dual variables associated with $\beta^1_{ij}(x^{ij}_i),\beta^2_{ij}(x^{ij}_j)$ respectively.  

We now re-write dual optimization in Eq \ref{eqn:dualForm}.  
\begin{align}
\label{reducVer}
\max_{\substack{\beta^0 \in \mathbb{R} \\ \beta^{1}_{ij}\geq 0 \\ \beta^2_{ij} \geq 0}} &
\beta^0 - \sum_{\substack{i \in \mathcal{Y}_p\\j \in \mathcal{Y}_q  }}  \beta^1_{ij}[x^{ij}_i=x^{y^*_p}_i]\\
\text{s.t.} \quad & -\lambda_{pq\rightarrow q}(y_q)- \beta^0 -  \sum_{\substack{i \in \mathcal{Y}_p\\j \in \mathcal{Y}_q }}   [x^{ij}_j=x_j^{y_q}]\beta^2_{ij} \geq 0 \quad \forall y_q \nonumber \\
                  & \theta_{ij} +\beta^1_{ij}+\beta^2_{ij}\geq 0 \quad \forall [i \in \mathcal{Y}_p, j \in \mathcal{Y}_q ]\nonumber
\end{align} 
Note that the objective in Eq \ref{reducVer} is non-increasing in $\beta^1$  thus $\beta^1$ is set to its minimum feasible value conditioned on $\beta^2$ at dual optimality. Therefor the following describes the relationship between $\beta^1_{ij} $ and $\beta^2_{ij}$.
\begin{align} 
\theta_{ij}+\beta^1_{ij}=-\beta^2_{ij}\\
\nonumber -\theta_{ij} \geq \beta^1_{ij}\geq 0
\end{align} 
We now rewrite dual optimization.  
\begin{align}
\max_{\substack{\beta^0 \in \mathbb{R} \\ -\theta_{ij} \geq \beta^{1}_{ij}\geq 0}} &
\beta^0 - \sum_{\substack{i \in \mathcal{Y}_p\\j \in \mathcal{Y}_q  }} \beta^1_{ij}[x^{ij}_i=x^{y^*_p}_i] \\
\text{s.t.} \quad & -\lambda_{pq\rightarrow q}(y_q) - \beta^0 + \sum_{\substack{i \in \mathcal{Y}_p\\j \in \mathcal{Y}_q }} [x^{ij}_j=x_j^{y_q}](\theta_{ij}+\beta^1_{ij}) \geq 0, \quad \forall y_q \nonumber 
\end{align} 
Notice that for $[x^{ij}_i=x^{y^*_p}_i] $ then the primal constraint corresponding to $\beta^1_{ij}$ is inactive since the constraint corresponding to $\beta^2_{ij}$ binds $\mu_{ij}(x_i^{ij},x_j^{ij})$ to be no greater than one.  We write this constraint  on $\beta^1_{ij}$ as  follows $-\theta_{ij} [x^{ij}_i \neq x^{y^*_p}_i]  \geq \beta^{1}_{ij}$.  We now rewrite dual optimization.  
\begin{align}
\max_{\substack{\beta^0 \in \mathbb{R} \\  \beta^{1}_{ij}\geq 0}} &
\beta^0 \\
\text{s.t.} \quad & -\lambda_{pq\rightarrow q}(y_q) - \beta^0 + \sum_{\substack{i \in \mathcal{Y}_p\\j \in \mathcal{Y}_q }} [x^{ij}_j=x_j^{y_q}](\theta_{ij}+\beta^1_{ij}) \geq 0 \quad \forall y_q \nonumber \\
&-\theta_{ij} [x^{ij}_i \neq x^{y^*_p}_i]  \geq \beta^1_{ij} \quad \quad \forall [i \in \mathcal{Y}_p,j \in \mathcal{Y}_q]\nonumber
\end{align} 
Observe that $\beta^0$ equals the primal optimization in Eq \ref{benders-gen} as $\beta^0$ is the only term in the dual objective and thus must equal the primal objective.   We write the formula for $\beta^0$ below for clarification.
\begin{align}
\beta^0=\min_{y_q} \phi_{pq}(y^*_p, y_q) - \lambda_{pq \rightarrow q}(y_q)
\end{align}
We now rewrite dual optimization.
\begin{align}
\label{befB3add}
\beta_0+\max_{\substack{ \beta^{1}_{ij}\geq 0}} &
0\\
\text{s.t.} \quad & -\lambda_{pq\rightarrow q}(y_q) - \beta^0 + \sum_{\substack{i \in \mathcal{Y}_p\\j \in \mathcal{Y}_q }}  [x^{ij}_j=x_j^{y_q}](\theta_{ij}+\beta^1_{ij}) \geq 0, \quad \forall y_q \nonumber \\
&-\theta_{ij} [x^{ij}_i \neq x^{y^*_p}_i]  \geq \beta^{1}_{ij} \quad \forall [i \in \mathcal{Y}_p,j \in \mathcal{Y}_q] \nonumber 
\end{align} 
We now alter the order of the summation in the constraint over $y_q$.
\begin{align}
\beta_0+\max_{\substack{ \beta^{1}_{ij}\geq 0}} &
0\\
\text{s.t.} \quad & -\lambda_{pq\rightarrow q}(y_q) - \beta^0 + \sum_{\substack{i \in \mathcal{Y}_p\\j \in \mathcal{Y}_q }}  [x^{ij}_j=x_j^{y_q}]\theta_{ij} +\sum_{\substack{j \in \mathcal{Y}_q \\ x_j}}[x^{y_q}=x_j](\sum_{i \in \mathcal{Y}_p} \beta^{1}_{ij} [x_j=x^{ij}_j]) \nonumber \\ 
&-\theta_{ij} [x^{ij}_i \neq x^{y^*_p}_i]  \geq \beta^{1}_{ij} \quad \forall [i \in \mathcal{Y}_p,j \in \mathcal{Y}_q] \nonumber 
\end{align} 

We now rewrite the Magnanti-Wong cut considered in Eq \ref{paretoQpLpForm}  using the notation of this section recalling that $-\theta_{ij} [x^{ij}_i \neq x^{y^*_p}_i]  \geq \beta^{1}_{ij}$.
 \begin{align}
- \epsilon \sum_{i \in \mathcal{Y}_p} \sum_{\substack{j \in \mathcal{Y}_p}}\beta^1_{ij}=- \epsilon \sum_{\substack{j \in \mathcal{Y}_p \\ x_j}} \sum_{i \in \mathcal{Y}_p} \beta^1_{ij}[x_j=x^{ij}_j]
\end{align}
We now  decrease the number of variables considered  by compacting $(\sum_{i \in \mathcal{Y}_p} \beta^{1}_{ij} [x_j=x^{ij}_j])$ to a single term $\beta^3_{j}(x_j)$ for each $j,x_j$.  We write the relationship between $\beta^1$ and $\beta^3$ below using constant terms $h_{ij},Q_j(x_j)$ .
\begin{align}
\beta^3_j(x_j)=\sum_{i \in \mathcal{Y}_p}\beta^1_{ij}[x_j=x^{ij}_j] \quad  \forall[j \in \mathcal{Y}_q,x_j]\\
\nonumber \beta^1_{ij}=\beta^3_{j}(x^{ij}_j)h_{ij}\\
\nonumber h_{ij}=\frac{-\theta_{ij} [x^{ij}_i \neq x^{y^*_p}_i] }{Q_j(x^{ij}_j)} \quad  \forall[i \in \mathcal{Y}_p,j \in \mathcal{Y}_q]\\
\nonumber Q_j(x_j)=\sum_{i \in \mathcal{Y}_p}-\theta_{ij} [x^{ij}_i \neq x^{y^*_p}_i] [x_j=x^{ij}_j] \quad  \forall[j \in \mathcal{Y}_q,x_j]
\end{align}
Using $\beta^3$ we now rewrite dual optimization.
\begin{align}
\beta_0+\epsilon \max_{\substack{Q_j(x_j)\geq \beta^{3}_{j}(x_j) \geq 0}} &
-\sum_{\substack{j \in \mathcal{Y}_q \\ x_j}}\beta^3_{j}(x_j)\\
\text{s.t.} \quad & -\lambda_{pq\rightarrow q}(y_q) - \beta^0 + \sum_{\substack{i \in \mathcal{Y}_p\\j \in \mathcal{Y}_q }} [x^{ij}_j=x_j^{y_q}](\theta_{ij}+\beta^3_{j}(x^{ij}_j)h_{ij}) \geq 0, \quad \forall y_q \nonumber 
\end{align} 
We observe empirically that using a mixture of the L1 and L2 norm over $\beta^3$ often improves performance of the entire procedure (Alg \ref{MPBasic}).  We write optimization parameterized by $\alpha \in [0,1]$ below. 

\begin{align}
\beta_0+\epsilon \max_{\substack{Q_j(x_j)\geq \beta^{3}_{j}(x_j) \geq 0}} &
- \sum_{\substack{j \in \mathcal{Y}_q\\ x_j}}\alpha\beta^3_{j}(x_j)+(1-\alpha)\beta^3_{j}(x_j)\beta^3_{j}(x_j)\\
\text{s.t.} \quad & -\lambda_{pq\rightarrow q}(y_q) - \beta^0 + \sum_{\substack{i \in \mathcal{Y}_p\\j \in \mathcal{Y}_q }} [x^{ij}_j=x_j^{y_q}](\theta_{ij}+\beta^3_{j}(x^{ij}_j)h_{ij}) \geq 0, \quad \forall y_q \nonumber 
\end{align} 
\section{Conversion Operation for Binary Problems}
\label{binarprop}
We consider a single cost table $\theta_{ij}(x_i,x_j)$ over binary variables $i,j$ and their states $x_i,x_j$.  In this section our goal is to alter $\theta_{ij}(x_i,x_j)$ so that at most one entry is non-zero and that entry is non-positive.  To do this we alter the unary terms $\theta_{i}(x_i)$ and  $\theta_{j}(x_j)$ so as to preserve the value of the objective across $x_i,x_j$.  Consider the functions in table form below.  
\begin{align}
\theta_{ij}(x_i,x_j)=
 \begin{bmatrix}
       \theta_{ij}(0,0),\theta_{ij}(0,1)\\ \theta_{ij}(1,0),\theta_{ij}(1,1)
    \end{bmatrix} \\
\nonumber    \theta_{i}(x_i)=
 \begin{bmatrix}
       \theta_{i}(0) \\  \theta_{i}(1)
    \end{bmatrix}\\
    \nonumber    \theta_{j}(x_j)=
 \begin{bmatrix}
       \theta_{j}(0) \\  \theta_{j}(1)
    \end{bmatrix}
%
    %
    \end{align}
    We now consider two cases.  In case one we let $\theta_{ij}(1,0)-\theta_{ij}(0,0)>\theta_{ij}(1,1)-\theta_{ij}(0,1)$ while in case two we assume that $\theta_{ij}(1,0)-\theta_{ij}(0,0)
    \leq \theta_{ij}(1,1)-\theta_{ij}(0,1)$.   In both cases we transform the tables by adding constants to rows, columns such that that the table $\theta_{ij}(x_i,x_j)$ contains exactly one non-zero entry and that entry is non-positive.  
    \subsection{Case One}
 We first subtract the element $\theta_{ij}(0,0)$ from the left column, $\theta_{ij}(0,1)$ from the right column,  and $\theta_{ij}(1,0)-\theta_{ij}(0,0)$ from the bottom row.  The corresponding elements are added to the unary tables.  This produces the following new tables.
 \begin{align}
 \theta_{ij}(x_i,x_j)=
  \begin{bmatrix}
      0,0\\ 0,\theta_{ij}(1,1)-\theta_{ij}(0,1)-\theta_{ij}(1,0) +\theta_{ij}(0,0)
    \end{bmatrix}\\
\nonumber    \theta_{i}(x_i)=
 \begin{bmatrix}
       \theta_{i}(0) +\theta_{ij}(0,0)\\  \theta_{i}(1)+\theta_{ij}(0,0)+\theta_{ij}(1,0) -\theta_{ij}(0,0)
    \end{bmatrix}\\
    \nonumber    \theta_{j}(x_j)=
 \begin{bmatrix}
       \theta_{j}(0) +\theta_{ij}(0,1)\\  \theta_{j}(1)+\theta_{ij}(0,1)+\theta_{ij}(1,0) -\theta_{ij}(0,0)
    \end{bmatrix}
 \end{align}
 Observe that the table $\theta_{ij}(x_i,x_j)$ contains exactly one non-zero entry and that entry is non-positive.  
    \subsection{Case Two}
 We first subtract the element $\theta_{ij}(0,0)$ from the left column and $\theta_{ij}(0,1)$ from the right column  and $\theta_{ij}(1,1)-\theta_{ij}(0,1)$ from the bottom row.  The corresponding elements are added to the unary tables.  This produces the following new tables.
 \begin{align}
 \theta_{ij}(x_i,x_j)=
  \begin{bmatrix}
      0,0\\ \theta_{ij}(1,0)-\theta_{ij}(0,0)-\theta_{ij}(1,1) +\theta_{ij}(0,1),0
    \end{bmatrix}\\
\nonumber    \theta_{i}(x_i)=
 \begin{bmatrix}
       \theta_{i}(0) +\theta_{ij}(0,0)\\  \theta_{i}(1)+\theta_{ij}(0,0)+\theta_{ij}(1,1) -\theta_{ij}(0,1)
    \end{bmatrix}\\
    \nonumber    \theta_{j}(x_j)=
 \begin{bmatrix}
       \theta_{j}(0) +\theta_{ij}(0,1)\\  \theta_{j}(1)+\theta_{ij}(0,1)+\theta_{ij}(1,1) -\theta_{ij}(0,1)
    \end{bmatrix}
 \end{align}
 Observe that the table $\theta_{ij}(x_i,x_j)$ contains exactly one non-zero entry and that entry is non-positive.  

\section{Derivation of Reverse Magnanti-Wong cuts}
\label{revDeriv}
Consider the dual LP for constructing Benders rows in Eq \ref{eqn:dualForm}.  Observe that for any $[i,j,x_i]$ such that $[x_i=x^{y^*_p}_i]=1$ that the primal constraint corresponding to $\beta^1_{ij}(x_i)$ is inactive since $\mu$ terms are upper bounded by one by the constraints over $\beta^2$ and $\beta^0$.  Thus we  rewrite optimization as follows. 
\begin{align}
\max_{\substack{\beta^0 \in \mathbb{R} \\ \beta^{1}_{ij}(x_i)\geq 0 \\ \beta^2_{ij}(x_j) \geq 0}} &
\beta^0 \\
\text{s.t.} \quad &-\lambda_{pq\rightarrow q}(y_q) - \beta^0 - \sum_{\substack{i \in \mathcal{Y}_p\\ j \in \mathcal{Y}_q \\x_j }}  [x_j=x_j^{y_q}]\beta^2_{ij}(x_j) \geq 0, \quad \forall y_q \nonumber \\
                  & \theta_{ij}(x_i,x_j) +\beta^1_{ij}(x_i)+\beta^2_{ij}(x_j)\geq 0, \quad \forall[ i \in \mathcal{Y}_p, j \in \mathcal{Y}_q,x_i,x_j] \nonumber\\
	& \beta^1_{ij}(x_i)=0 \quad \forall [ i \in \mathcal{Y}_p, j \in \mathcal{Y}_q,x_i=x^{y^*_p}_i] \nonumber
\end{align} 
Since $\beta^0$ is the only term in the objective it must equal the optimal primal objective at dual optimality.

%
%
Observe that larger values of $\beta^2$ make the constraints over $y_q$ more difficult to satisfy.  Thus at dual optimality $\beta^2$ is set to the smallest feasible value given $\beta^1$ which we write below.  
\begin{align}
\beta^2_{ij}(x_j)=\max(0,-\min_{x_i} \theta_{ij}(x_i,x_j) +\beta^1_{ij}(x_i) ) \quad \quad \forall [i \in \mathcal{Y}_p,j \in \mathcal{Y}_q, x_j]
\end{align}
%
%
The smallest possible value of $\beta^2$ is achieved by setting $\beta^1$ as follows.  
\begin{align}
\beta^1_{ij}(x_i )= -[x_i \neq x^{y^*_p}_i](\min_{x_j} \theta_{ij}(x_i,x_j)) \quad \quad \forall [i \in \mathcal{Y}_p,j \in \mathcal{Y}_q, x_i]
\end{align}
Since $\beta^2$ is set to its element-wise minimum value and the dual objective equals  the primal objective then the solution below must be feasible.   
\begin{itemize}
\item
$\beta^0$ is set equal to $ \min_{y_q} \phi_{pq}(y^*_p, y_q) - \lambda_{pq \rightarrow q}(y_q)$.
\item
 $\beta^1_{ij}(x_i)\leftarrow -[x_i \neq x_i^{y_p^*}]\min_{x_j}\theta_{ij}(x_i,x_j)$ for all $i\in \mathcal{Y}_p,j\in \mathcal{Y}_q,x_i$.  
\item 
 $\beta^2_{ij}(x_j)\leftarrow -\min_{x_i}\theta_{ij}(x_i,x_j)+\beta^1_{ij}(x_i)$ for all $i\in \mathcal{Y}_p,j\in \mathcal{Y}_q,x_j$.  
\end{itemize}
%
%
%
\end{document}